\documentclass[letterpaper, 10 pt, conference]{ieeeconf}  %

\IEEEoverridecommandlockouts                              %

\usepackage{cite}
\usepackage{lipsum}
\usepackage[inkscapelatex=false]{svg}
\usepackage{amssymb,amsfonts}
\usepackage[fleqn]{amsmath}
\usepackage{algorithmic}
\usepackage{booktabs}
\usepackage{graphicx}
\usepackage{pifont}
\usepackage{textcomp}
\usepackage{xcolor} 
\usepackage{array}
\usepackage{cellspace}
\usepackage{array}
\usepackage{colortbl} %
\usepackage{multirow}
\usepackage{subcaption}
\usepackage{color,soul}
\usepackage{siunitx}   %
\usepackage{adjustbox}

\def\BibTeX{{\rm B\kern-.05em{\sc i\kern-.025em b}\kern-.08em
    T\kern-.1667em\lower.7ex\hbox{E}\kern-.125emX}}

\newsavebox\CBox
\def\textBF#1{\sbox\CBox{#1}\resizebox{\wd\CBox}{\ht\CBox}{\textbf{#1}}}

\begin{document}

\title{
Discrete Contrastive Learning for Diffusion Policies in Autonomous Driving}

\author{Kalle Kujanp\"a\"a$^{1,3}$, Daulet Baimukashev$^{2}$, Farzeen Munir$^{2,3}$, Shoaib Azam$^{2,3}$, \\
Tomasz Piotr Kucner$^{2,3}$, Joni Pajarinen$^{2,3}$, and Ville Kyrki$^{2,3}$%
\thanks{$^{1}$Department of Computer Science, Aalto University, Finland.}
\thanks{$^{2}$Department of Electrical Engineering and Automation, Aalto University}
\thanks{$^{3}$Finnish Center for Artificial Intelligence (FCAI)}
\thanks{Corresponding author: Shoaib Azam (shoaib.azam@aalto.fi)}
\thanks{All emails \{first.last\}@aalto.fi. We acknowledge the computational resources provided by the Aalto Science-IT project and CSC, Finnish IT Center for Science. The work was funded by Research Council of Finland (aka Academy of Finland) within the Flagship Programme, Finnish Center for Artificial Intelligence (FCAI). J.~Pajarinen was partly supported by Research Council of Finland (aka Academy of Finland) (345521).}%
}

\maketitle

\begin{abstract}
Learning to perform accurate and rich simulations of human driving behaviors from data for autonomous vehicle testing remains challenging due to human driving styles' high diversity and variance. We address this challenge by proposing a novel approach that leverages contrastive learning to extract a dictionary of driving styles from pre-existing human driving data. We discretize these styles with quantization, and the styles are used to learn a conditional diffusion policy for simulating human drivers. Our empirical evaluation confirms that the behaviors generated by our approach are both safer and more human-like than those of the machine-learning-based baseline methods. We believe this has the potential to enable higher realism and more effective techniques for evaluating and improving the performance of autonomous vehicles.
\end{abstract}

\section{Introduction}

Human behavior modeling is essential in developing intelligent learning systems, particularly for autonomous driving. This approach offers crucial insights into how humans interact, behave, and respond within complex environments \cite{feng2021intelligent}. However, modeling and simulating human behavior to create a data-driven simulation for autonomous driving faces challenges due to the high diversity and variability of human driving behaviors \cite{fuchs2022modeling, chen2023data}. Accurately capturing these behaviors is critical for enhancing simulation realism, which in turn improves sim-to-real transfer and facilitates safer autonomous vehicle development.
\par
Imitation learning approaches, particularly Behavior Cloning (BC), offer a viable method to model human behaviors \cite{zheng2022imitation}. In BC, an agent is trained to emulate demonstrator actions using an offline dataset of observation-action pairs, effectively replicating human-like decision-making processes. However, popular BC approaches often simplify distribution modeling for accessible learning, typically using point estimates via Mean Squared Error (MSE) or action space discretization \cite{ke2021imitation,urain2024deep}. These limitations highlight the need for an improved behavior model that can capture long and short-term dependencies, represent diverse modes of behavior, and accurately replicate learned behaviors.
\par
Recent advancements in diffusion models, currently leading in image, video, and audio generation \cite{saharia2022photorealistic, harvey2022flexible, kong2020diffwave, huang2023noise2music, ho2022video}, offer a promising approach to modeling the full spectrum of human driving behavior. Prior work has demonstrated how to model action distributions without approximation with diffusion, capturing the complete action distribution \cite{pearce2023imitating}. This could overcome the limitations of traditional behavior modeling techniques in autonomous driving, but it may struggle to fully capture the diversity inherent in human driving due to not explicitly modeling how temporally consistent human driving styles affect driving behavior.
\begin{figure}
\centering
\includegraphics[width=\columnwidth]{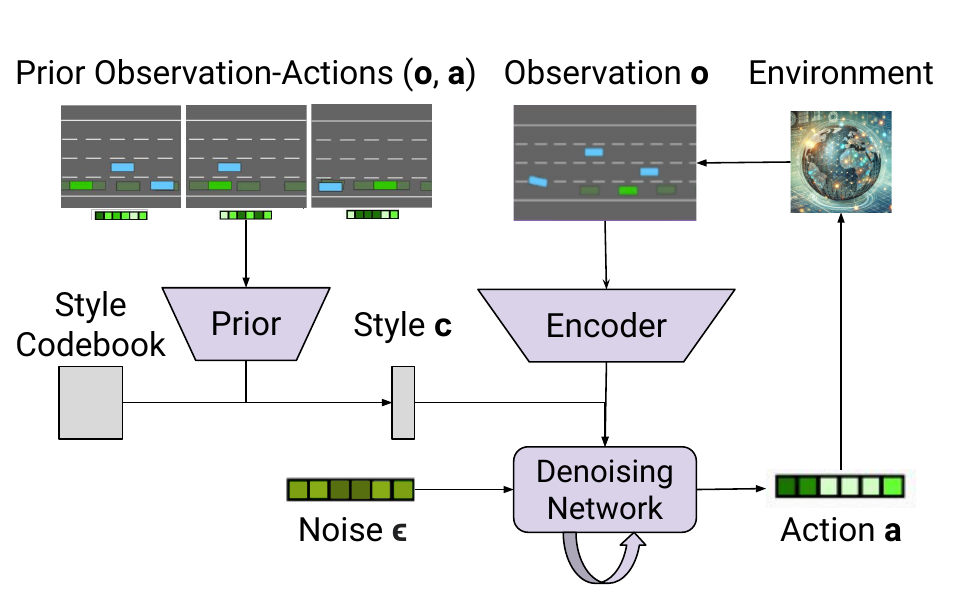}
\caption{An overview of our method, DSDP (Discrete Style Diffusion Policy). DSDP takes a history of past driving behavior as observation-action pairs $(\mathbf{o}, \mathbf{a})$ and uses a prior network to sample a driving style $\mathbf{c}$. The styles are learned via discrete contrastive learning. The denoising network then uses the encoded current observation $\mathbf{o}$ and style $\mathbf{c}$ to generate a driving action $\mathbf{a}$ through diffusion.}
\label{fig:overview}
\end{figure}

In this work, we propose \textbf{DSDP: Discrete Style Diffusion Policy}, a novel framework for learning to model human driving behavior from pre-existing datasets. Our approach is predicated on the assumption that human drivers exhibit a variety of distinct driving styles, contributing to the diversity of observed driving behaviors. To capture this diversity, DSDP employs contrastive learning with the InfoNCE loss \cite{oord2018representation} and Lookup-Free Quantization (LFQ) \cite{yu2023language} to extract discrete driving styles from unlabeled data. Then, we train a conditional Denoising Diffusion Probabilistic Model (DDPM) \cite{ho2020denoising} conditioned on both observations and driving styles to generate actions. 
By integrating these components, DSDP aims to overcome the limitations of traditional behavior cloning techniques in representing diverse human driving behaviors. We evaluate the performance of our proposed method in various environments, demonstrating superior performance in capturing the range and fidelity of human driving behavior compared to the baselines and parametric models. Our results suggest that incorporating explicit driving styles can improve the realism of driving simulations, which is critical for training and validating autonomous vehicle policies.
\par
The main contributions of our work are:
\begin{enumerate}
  \item An unsupervised discrete driving style extraction method using the contrastive InfoNCE loss and LFQ on unlabeled data to model human driving behavior.
  \item DSDP, a method that combines the driving style extraction with a conditional DDPM for action generation
  \item A comprehensive evaluation on historical driving data with a series of ablation studies
\end{enumerate}

\section{Related Work}

Human driving behavior is complex, involving interactions between the driver, vehicle, and environment and integrating various psychological aspects, complicating modeling \cite{brown2020modeling, negash2023driver}. The variability and diversity of drivers further complicate developing Driver Behavior Models (DBMs) \cite{negash2023driver}. Traditional DBMs like the Intelligent Driver Model (IDM) \cite{treiber2000congested}, MOBIL \cite{kesting2007general}, and models by Gipps \cite{gipps1981behavioural, gipps1986model}, are primarily deterministic. Despite attempts to calibrate them with real-world data \cite{sangster2013application,li2016global,punzo2012can,ma2020sequence}, they often fail to capture the stochastic nature of human driving behavior.

Recent research has increasingly turned to machine learning, particularly neural networks, to fit DBMs to large driving datasets \cite{xie2019data,huang2018car,wang2017capturing}. However, even these approaches struggle to capture the inherent stochasticity of driving behavior. While adding noise can introduce randomness, simple approaches like Gaussian noise often fail to capture variability in human driving behavior \cite{laval2014parsimonious,treiber2017intelligent, yang2010development, kuefler2017imitating}. Stochasticity can also be introduced through model parameters, such as by using game theory \cite{talebpour2015modeling} or genetic algorithms \cite{hamdar2015behavioral}. \cite{wang2009markov} and \cite{chen2010markov} proposed stochastic car-following models that effectively represent the distribution of time progress to align simulation environments more closely with real-world data. However, these studies focus primarily on single-lane road scenarios and do not address the accumulation of errors over time.

BC models human behavior by directly learning policies from demonstrating data, minimizing the forward loss relative to the expert distribution \cite{ghasemipour2020divergence,osa2018algorithmic}. However, naïve BC struggles with multimodal distributions often present in real-world datasets from multiple demonstrators \cite{urain2024deep}. To overcome this, advanced models like energy-based \cite{florence2022implicit} and generative adversarial models \cite{ho2016generative} have been explored. However, energy-based models suffer from complex sampling, while generative adversarial models face mode collapse \cite{srivastava2017veegan}, limiting their suitability for modeling driving behavior.

Recently, diffusion models have gained popularity for policy learning, demonstrating success in, among others, robotic control and 3D gaming environments \cite{wang2022diffusion,ajay2022conditional,hansen2023idql,reuss2023goal,pearce2023imitating}. Diffusion models effectively capture multimodal, complex distributions and offer more straightforward sampling compared to energy-based models \cite{florence2022implicit}. Planning with diffusion models \cite{he2024diffusion,mishra2023generative,janner2022planning} is an alternative to using them to represent the BC policy. In contrast, consistency models offer a faster, one-step sampling alternative to diffusion models with possibly lower generation quality \cite{ding2023consistency, song2023consistency}. We demonstrate that modeling human driving style explicitly while using diffusion as the policy learning backbone achieves more accurate results than directly using diffusion.

\begin{figure*}
\centering
\includegraphics[width=\textwidth]{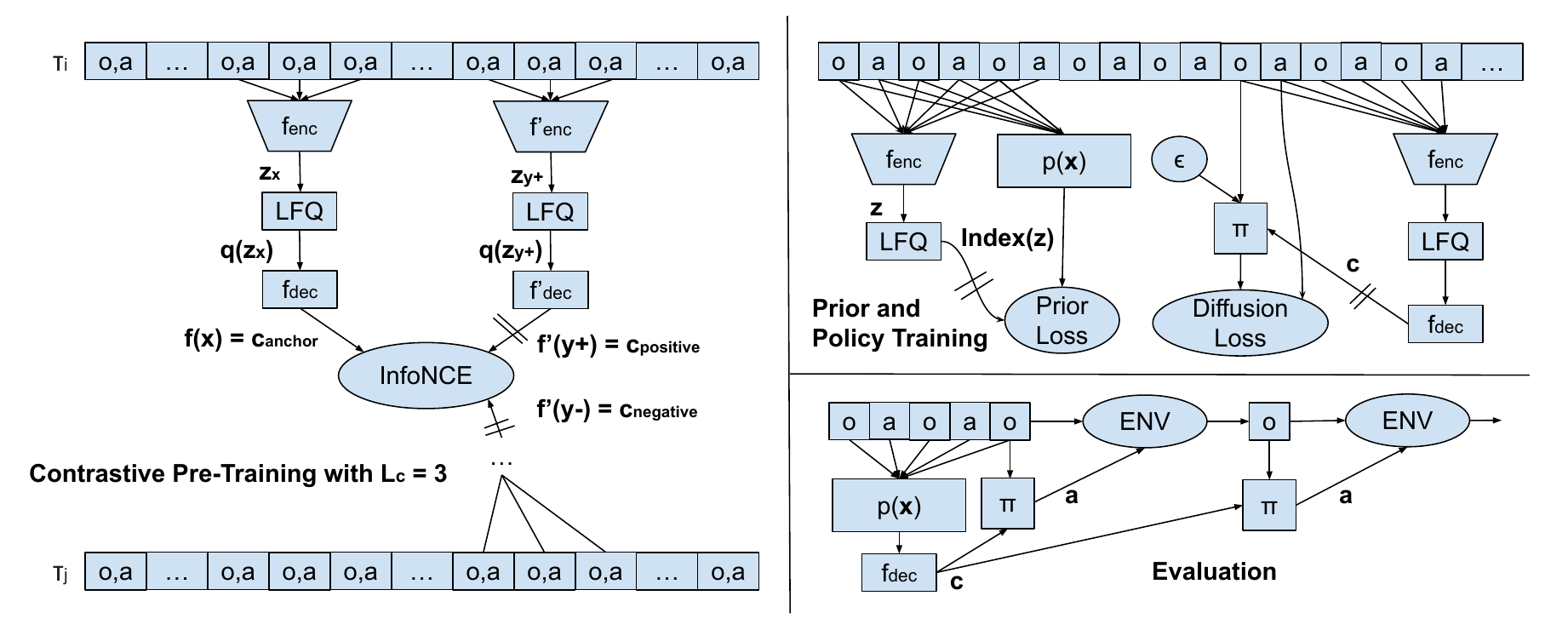}
\caption{Contrastive pre-training (left), policy and prior training (upper right), and the evaluation (bottom right). During contrastive pre-training, we sample two sub-trajectories $\mathbf{x} = (\mathbf{o}_{i,1}, \mathbf{a}_{i,1}, \dots, \mathbf{o}_{i,L_c}, \mathbf{a}_{i,L_c})$ and $\mathbf{y+}$ from each trajectory $\tau_i$, process them with the contrastive representation function consisting of the encoder $f_\text{enc}$, the LFQ discretization layer and the decoder $f_\text{dec}$ to get the styles $\mathbf{c}_\text{anchor}, \mathbf{c}_\text{positive}$. We utilize the styles corresponding to the sub-trajectories from other trajectories in the batch as negatives, $\mathbf{c}_\text{negative}$ in the computation of the InfoNCE loss. The prior $p(\mathbf{x})$ is trained to predict the index of the encoded sub-trajectory $\mathbf{x}$, and the policy $\pi$ to minimize the DDPM loss. Two parallel lines signify cutting the gradient flow; the contrastive representation function f is fixed during prior and policy training.} 
\label{fig:method}
\end{figure*}

\section{Method}

This section introduces our BC setting and proposes a novel contrastive learning approach for unsupervised driving style identification, followed by discretization for efficient sampling of driving styles. Finally, we integrate this with diffusion-based policy learning to formulate the proposed method, Discrete Style Diffusion Policy (DSDP).

\subsection{Behavior Cloning Setting}

We assume a BC setting to learn a policy that replicates human driving behaviors from a static dataset. Our goal is to model the distribution of human actions $\mathbf{a}$ given observations $\mathbf{o}$ with a policy $\pi(\mathbf{a} | \mathbf{o})$. The objective is to maximize driving success and align the learned policy's state visitation with the human policy. We use a dataset $\mathcal{D}$ with $N$ observation-action trajectories, $\mathcal{D} = \{\tau_0, \dots, \tau_N\}$, where each trajectory $\tau_i$ consists of paired observations and actions. The environment is partially observable, stochastic, and without rewards.

\subsection{Contrastive Learning of Driving Styles}

We aim to learn diverse driving styles from the dataset  $\mathcal{D}$ without prior assumptions or labeled style data. Given a style $\mathbf{c}$, we train a policy $\pi(\mathbf{a} | \mathbf{o}, \mathbf{c})$ to generate temporally consistent behaviors by fixing $\mathbf{c}$ during evaluation (see Fig.~\ref{fig:method}). Our approach uses contrastive learning, an approach that extracts meaningful representations from high-dimensional data and has been widely used for images \cite{park2020contrastive}, video \cite{wang2022long}, reinforcement learning \cite{zheng2024texttt}, and audio \cite{saeed2021contrastive}. Its core principle is that representations of different views of the same object should be similar, whereas representations of different objects should diverge. It emphasizes global structures while minimizing the influence of low-level details and noise, grouping positive pairs and separating negative pairs in the representation space \cite{oord2018representation}.

We use the InfoNCE loss function with representation function $f_\phi$, modeled by a neural network. Given $x \sim p(x)$, a corresponding positive sample $y_{+} \sim p(y | x)$ and a set of $N-1$ negative samples, $\mathcal{Y} = \{y_1, \dots, y_{N-1} \}$ drawn from $p(y)$, the InfoNCE loss is defined as:
\begin{equation}
    \mathcal{L}_\text{INFO} = -\mathbb{E}_{x \sim p(x)} \left[\log \frac{f_\phi(x, y_{+})}{\sum_{y \in \mathcal{Y} \cup \{y_{+} \}} f_\phi(x,y)} \right]
    \label{eq:infonce}
\end{equation}
Minimizing this loss with respect to the parameters $\phi$ maximizes the similarity between the representations of the positive pair $(x, y_{+})$ while pushing the representations of the negative samples away \cite{oord2018representation}. 

In our method, given a batch of $N$ driving trajectories $\{ \tau_1, \dots, \tau_N \}$, we sample two non-overlapping sub-trajectories, $x$ and $y_{+}$, each of length $L_c$ (see Fig.\ref{fig:method}, left) from each trajectory. Positive pairs come from the same trajectory, while negatives come from different trajectories in the same batch. The representation function $f_\phi(x) = \mathbf{c}_\text{anchor}$ captures the driver's driving style by filtering irrelevant details. To stabilize training and mitigate overfitting, the second sub-trajectory $y_{+}$ is processed through a target network $f'_{\phi'}(y_{+})$, whose parameters $\phi'$ are updated as a moving average of $\phi$. This approach effectively learns a function capable of extracting driving styles from the trajectories.

\subsection{Discretizing the Styles with LFQ}

The output of the representation function $f_\phi$ is typically continuous and high-dimensional, making it challenging to sample realistic driving styles due to the curse of dimensionality. We address this by discretizing the driving style representations using vector quantization, compressing the space into a finite set of styles, preserving key features and enabling efficient sampling.
Vector-quantized autoencoders (VQVAEs) are popular for learning discrete representations \cite{van2017neural}, but training can be challenging \cite{lancucki2020robust} and the final VQVAE can suffer from a disconnect between reconstruction and generation quality, which has prompted the development of alternatives \cite{yu2023language,esser2021taming,mentzer2023finite}.

We adopt Lookup-Free Quantization \cite{yu2023language}, which addresses these challenges by eliminating the explicit codebook lookup. In VQVAEs, the quantization layer discretizes representations by selecting the nearest vector from a codebook. LFQ replaces the codebook with an integer set $\mathbb{C}$, where $|\mathbb{C}| = K$. This approach avoids large embedding table lookups and simplifies the quantization process.
The latent space of LFQ is the Cartesian product of single-dimensional variables, $\mathbb{C} = \times_{i=1}^{\log_2 K} C_i$. For a continuous vector $\mathbf{z} \in \mathbb{R}^{log_2 K}$, quantization is performed dimension-wise:
\begin{equation}
    q(z_i) = C_{i, j}, \text{where } j = \arg \min_k || z_i - C_{i, k} ||_2,
\end{equation} where $C_{i, j}$ is the j-th value of $C_i$. When $C_i = \{-1, 1\}$, this simplifies to
\begin{equation}
    q(z_i) = \text{sign}(z_i) = -1\{z_i \le 0\} + 1\{z_i > 0\}.
\end{equation}
This discrete representation can be mapped to a specific integer index, $\text{Index}(\mathbf{z})$. To ensure that all potential values of $q(\mathbf{z})$ are utilized, we add an entropy penalty to the objective function \cite{yu2023language}. In addition to solving the issue of decoupled reconstruction and generation quality, we find that LFQs are easier to train than VQVAEs, with more stable learning dynamics and faster convergence.

The InfoNCE representation network $f_\phi$ presented in the previous subsection consists of an encoder, an LFQ discretization layer, and a decoder (see Fig. \ref{fig:method}, left). Unlike standard VQVAE training, which relies on a reconstruction loss, we utilize the InfoNCE loss supplemented with the LFQ entropy penalty. The encoder $f_{enc}(\mathbf{o}_i, \mathbf{a}_i, \dots, \mathbf{o}_{i+L_c-1}, \mathbf{a}_{i+L_c-1}) = \mathbf{z}$ processes a sequence of observation-action pairs into a continuous latent representation $\mathbf{z}$. The LFQ discretization layer quantizes $\mathbf{z}$ into a discrete code $q(\mathbf{z})$, serving as a bottleneck that compresses the high-dimensional continuous space into a finite set of discrete driving styles, facilitating efficient sampling. Finally, the decoder $f_\text{dec}$ maps the discrete representation $q(\mathbf{z})$ to a continuous style vector $\mathbf{c}$ for the InfoNCE loss.

At evaluation, the discrete styles $q(\mathbf{z})$ can be directly sampled and decoded to retrieve the driving style vectors $\mathbf{c}$. To sample the style, we learn a prior distribution $p(\text{Index}(\mathbf{z}) | \mathbf{o}_i, \mathbf{a}_i, \dots, \mathbf{o}_{i+L_p-1})$, which predicts the index of the style conditioned on the trajectory. The length of the sub-trajectory considered by the prior, $L_p$ can be different from the sub-trajectory length of the contrastive learning process, $L_c$. Training the prior is framed as a classification task with the cross-entropy loss (see Fig. \ref{fig:method}, top right).

\subsection{Diffusion Policy for BC}

In BC policy learning, a diffusion model starts from noise $\mathbf{a}_T \sim \mathcal{N}(\textbf{0}, \textbf{I})$, and iteratively produces a sequence of less noisy actions $\mathbf{a}_{T-1}, \dots, \mathbf{a}_0$, where $\mathbf{a}_0$ corresponds to the action $\mathbf{a}$. Normally, the denoising process is conditioned on observations $\mathbf{o}$ \cite{ho2020denoising, dhariwal2021diffusion, chi2023diffusion}, but we also condition on the driving style $\mathbf{c}$. We follow the Denoising Diffusion Probabilistic Model (DDPM) formulation \cite{ho2020denoising} to train a denoising network $\epsilon_\theta$ to minimize
\begin{equation}
     \mathcal{L}_\text{DDPM}(\theta) = \mathbb{E}_{\mathbf{o}, \mathbf{a} \sim \mathcal{D}, t \sim \mathcal{U}[1, T]} [|| \mathbf{\epsilon} - \epsilon_\theta(\mathbf{o}, \mathbf{a_t}, t, \mathbf{c}) ||_2^2],
\end{equation}
where $\epsilon \sim \mathcal{N}(\mathbf{0}, \mathbf{I})$ is random noise, $(\mathbf{o}, \mathbf{a})$ is sampled from the dataset $\mathcal{D}$, the denoising timestep $t$ is uniformly drawn from $[1, T]$, and the noisy action $\mathbf{a}_t$ is computed as $\mathbf{a}_t = \sqrt{\bar{\alpha_t}}\mathbf{a} + \sqrt{1 - \bar{\alpha_t}}\mathbf{\epsilon}$. $\bar{\alpha_t}$ comes from a pre-determined noise schedule \cite{ho2020denoising}. The driving style $\mathbf{c}$ is derived from the representation function $f_\phi$ taking the ground-truth sub-trajectory data as input, which makes the learning task easier and minimizes the probability of the policy ignoring the style (see Fig.~\ref{fig:method}, top right).

During sampling, DDPM starts from random noise $\mathbf{a}_T$ and performs T denoising steps:
\begin{equation}
    \mathbf{a}_{t-1} = \frac{1}{\sqrt{\alpha_t}}(\mathbf{a}_t - \frac{1-\alpha_t}{\sqrt{1-\bar{\alpha_t}}}\mathbf{\epsilon}_\theta(\mathbf{o}, \mathbf{a}_t, t, \mathbf{c})) + \sigma_t \epsilon,
\end{equation}
where $\epsilon \sim \mathcal{N}(\mathbf{0}, \mathbf{I})$ unless $t=1$ \cite{ho2020denoising}. We use the standard DDPM noise schedule \cite{ho2020denoising}. The style $\mathbf{c}$ is determined by the prior that receives the observation $\mathbf{o}$ and its preceding $L_p-1$ observation-action pairs as input. We sample an index from the resulting logits and pass the corresponding latent variable through the decoder $f_\text{dec}$. The sampled style remains fixed throughout the rest of the trajectory. Unlike during training, the policy cannot access the style computed from the future ground-truth actions (see Fig.~\ref{fig:method}, bottom right).

\section{Experiments}

\begin{table*}
\caption{Crashing percentage on our four benchmark environments across five seeds. Our method, DSDP, exhibits the best performance among fully learning-based methods, has the lowest crashing percentage in three environments, and is within the uncertainty in the fourth one. The uncertainty is two standard errors.}
\centering
\begin{tabular}{l *{5}{S[table-format=1.1] S[table-format=2.2]}}
\hline
Method & \multicolumn{2}{c}{US 101} & \multicolumn{2}{c}{I-80} & \multicolumn{2}{c}{Lankershim} & \multicolumn{2}{c}{Peachtree} & \multicolumn{2}{c}{Overall} \\ \hline
DSDP (ours) & \textbf{0.2} &\pm0.36 & \textbf{3.0} &\pm2.19 & \textbf{8.6} &\pm2.09 & 4.4 &\pm0.91 & \textbf{4.0} & \pm0.79 \\
Diffusion-BC & 0.5 &\pm0.50 & 7.0 &\pm0.57 & 13.0 &\pm0.80 & 4.7 &\pm0.54 & 6.3 & \pm0.31 \\
Uncond. Diffusion-BC & 0.4 &\pm0.44 & 7.8 &\pm0.88 & 13.6 &\pm0.72 & \textbf{3.7} &\pm0.54 & 6.4 & \pm0.33 \\
MSE & 0.8 &\pm0.36 & 8.8 &\pm0.67 & 12.6 &\pm0.72 & 3.8 &\pm0.67 & 6.5 & \pm0.31 \\
Discretised & 1.2 &\pm0.36 & 10.2 &\pm0.67 & 16.4 &\pm0.44 & 6.6 &\pm0.44 & 8.6 & \pm0.24 \\
Gaussian & 0.4 &\pm0.44 & 8.5 &\pm1.12 & 13.8 &\pm0.36 & 4.8 &\pm0.36 & 6.9 & \pm0.33 \\
K-Means & 1.0 &\pm0.00 & 9.2 &\pm1.43 & 14.4 &\pm0.44 & 5.0 &\pm0.57 & 7.4 & \pm0.40 \\
K-Means+Residual & 0.6 &\pm0.44 & 8.6 &\pm0.72 & 14.2 &\pm0.36 & 4.6 &\pm0.72 & 7.0 & \pm0.29 \\
EBM Deriv-Free & 10.0 &\pm17.44 & 5.6 &\pm3.60 & 12.4 &\pm4.26 & 4.8 &\pm2.55 & 8.2 & \pm4.62 \\
\hline
IDM (Fixed) & 0.0 &\pm0.00 & 0.0 &\pm0.00 & 0.0 &\pm0.00 & 0.0 &\pm0.00 & 0.0 & \pm0.00 \\
IDM (Learned) & 0.0 &\pm0.00 & 0.0 &\pm0.00 & 0.0 &\pm0.00 & 0.0 &\pm0.00 & 0.0 & \pm0.00 \\
\hline
\end{tabular}
\label{tab:highway_env_crash_prob}
\end{table*}

\begin{table*}
\caption{The F1 Score measures the similarity between generated trajectories and ground-truth trajectories on four benchmark environments across five seeds. Our method shows the best performance overall. The uncertainty is two standard errors. Even the learning-based IDM fails to match the human trajectories.}
\centering
\begin{tabular}{l *{5}{S[table-format=1.3] S[table-format=1.2]}}
\hline
Method & \multicolumn{2}{c}{US 101} & \multicolumn{2}{c}{I-80} & \multicolumn{2}{c}{Lankershim} & \multicolumn{2}{c}{Peachtree} & \multicolumn{2}{c}{Overall} \\ \hline
DSDP (ours) & \textbf{0.539} &\pm0.00 & \textbf{0.464} &\pm0.01 & \textbf{0.367} &\pm0.02 & 0.244 &\pm0.01 & \textbf{0.404} & \pm0.01 \\
Diffusion-BC & 0.486 &\pm0.01 & 0.425 &\pm0.01 & 0.340 &\pm0.00 & 0.254 &\pm0.01 & 0.376 & \pm0.00 \\
Uncond. Diffusion-BC & 0.470 &\pm0.00 & 0.428 &\pm0.00 & 0.343 &\pm0.01 & 0.253 &\pm0.02 & 0.374 & \pm0.01 \\
MSE & 0.479 &\pm0.00 & 0.425 &\pm0.01 & 0.326 &\pm0.00 & 0.251 &\pm0.01 & 0.370 & \pm0.00 \\
Discretised & 0.483 &\pm0.01 & \textbf{0.464} &\pm0.01 & 0.344 &\pm0.01 & 0.255 &\pm0.01 & 0.387 & \pm0.00 \\
Gaussian & 0.498 &\pm0.01 & \textbf{0.464} &\pm0.00 & 0.333 &\pm0.01 & \textbf{0.258} &\pm0.00 & 0.388 & \pm0.00 \\
K-Means & 0.499 &\pm0.01 & 0.461 &\pm0.00 & 0.339 &\pm0.01 & 0.254 &\pm0.01 & 0.388 & \pm0.00 \\
K-Means+Residual & 0.511 &\pm0.01 & 0.463 &\pm0.01 & 0.335 &\pm0.00 & 0.257 &\pm0.00 & 0.391 & \pm0.00 \\
EBM Deriv-Free & 0.243 &\pm0.10 & 0.341 &\pm0.04 & 0.291 &\pm0.04 & 0.227 &\pm0.03 & 0.276 & \pm0.03 \\
\hline
IDM (Fixed) & 0.030 &\pm0.00 & 0.034 &\pm0.00 & 0.064 &\pm0.04 & 0.051 &\pm0.01 & 0.045 & \pm0.01 \\
IDM (Learned) & 0.030 &\pm0.00 & 0.034 &\pm0.00 & 0.064 &\pm0.04 & 0.051 &\pm0.01 & 0.045 & \pm0.01 \\
\hline
\end{tabular}
\label{tab:multistep_f1}
\end{table*}

Our evaluation of the Discrete Style Diffusion Policy (DSDP) focuses on three key aspects: 1) \textbf{Safety:} We assess the model's ability to exhibit safe driving behaviors by comparing crash percentages with baseline methods. 2) \textbf{Human-likeness:} We evaluate how well the model captures human driving behavior by comparing generated trajectories to ground-truth data using density and coverage metrics combined into an F1 score \cite{naeem2020reliable}. 3) \textbf{Design choices:} We conduct ablation studies to understand the contribution of different components of our method, including sampling strategies, architectural choices, and hyperparameters.

\subsection{Evaluation with NGSIM}

We evaluate our approach using the Next Generation Simulation (NGSIM) dataset \cite{ngsim}, consisting of trajectories from thousands of drivers across four U.S. locations - US 101, I-80, Lankershim Boulevard, and Peachtree Street and containing extensive human driving patterns. The dataset records trajectories at 100 ms intervals, including vehicle position, lane number, velocity, acceleration, vehicle type, dimensions, and IDs of preceding and following vehicles. We use the NGSIM dataset for a car-following setting, the standard use case for IDM. IDM is still widely used in the setting, even though it can generate unhumanlike behaviors.

We preprocess the NGSIM dataset by normalizing the data between 0 and 1. For vehicles without a preceding vehicle, we set the space headway to the dataset's maximum value. We introduce a new time headway variable by dividing the space headway by the current velocity to complement the noisy original. To make the state as Markovian as possible, we add the current and previous velocity of the preceding vehicle. We cap speed changes to gravity per second. We discarded outliers, where the speed changes drastically for at most 200 ms before returning the baseline. Finally, we apply Savitzky-Golay smoothing \cite{savitzky1964smoothing} to the dataset. The action to be reconstructed is the vehicle acceleration. We split the dataset chronologically: the first 80 \% for training and the remaining 20 \% for testing. Trajectories are sampled with $L_p = 5$, corresponding to observing the first 0.5 seconds before simulation with the learned policy.

We use convolutional networks for the prior $p$ and encoder $f_{enc}$, and a standard MLP as the decoder $f_{dec}$. The denoising network $\epsilon_\theta$ is an MLP sieve network \cite{pearce2023imitating}, which encodes each variable $\mathbf{o}, \mathbf{a}_t, \mathbf{t}$ separately. During training, the representation function $f_\phi$ extracts the style $\mathbf{c}$, which is concatenated with the embeddings as input to the denoising MLP. The MLP uses skip connections, with embeddings of the denoising timestep $t$ and the noisy action $\mathbf{a}_t$ concatenated after each hidden layer.
We train the methods for 30 epochs, taking a checkpoint every five epochs and reporting the best checkpoint's performance. For the diffusion model, we use a batch size of 32, a learning rate of 0.0001, and a hidden dimension of 128. Contrastive pre-training is performed over 500 passes with a batch size of 128 and a learning rate of 0.001, sampling one sub-trajectory pair per trajectory per pass. The model uses 16 channels, a codebook of size 256, and a 64-dimensional style variable $\mathbf{c}$.

The F1 score is calculated by sampling traffic scenarios from the NGSIM test dataset, randomly selecting one vehicle as the ego vehicle while keeping the other vehicles' behaviors fixed based on historical data. The driving policy $\pi(\mathbf{a} | \mathbf{o}, \mathbf{c})$ determines the actions of the ego vehicle. We compute density and coverage \cite{naeem2020reliable} using the sampled observations, with the historical behavior of the ego vehicle serving as ground truth. The F1 score is the harmonic mean of density and coverage.

\subsection{Evaluation in Highway-ENV}

We evaluate safe driving behavior using highway-env \cite{highway}. We recreate the four NGSIM road scenarios. To make the simulations reactive, we implement an IDM policy for the other vehicles instead of log-replayed trajectories. This prevents crashes due to incorrect transitions in the NGSIM data, such as stationary vehicles with non-zero velocities. At evaluation, the ego vehicle is randomly sampled from the NGSIM data, with initial conditions (position, velocity, lane ID) set from the data. The preceding vehicle's target speed is set to its maximum recorded speed. Each scenario is run for 200 timesteps at 10 Hz. A crash is counted if the ego vehicle collides within this period. We repeat this for 100 sampled ego vehicles and calculate the crash percentage.

\subsection{Baselines}

We compare DSDP to standard diffusion policies learned with DDPM \cite{ho2020denoising, pearce2023imitating}. Since our model has access to the first $L_p$ timesteps (0.5 seconds of the trajectory), we add a conditioning network to the baseline policy, which processes the initial trajectory and concatenates the representation with the observation. This network is trained end-to-end, and we also evaluate the diffusion policies without it.

Other baselines include a standard BC policy $p(\mathbf{a} | \mathbf{o})$ trained with a mean-squared error objective, which outputs point estimates lacking variability and multimodality. Another baseline discretizes actions into $B=20$ bins and treats the task as classification. This approach can model variability and multimodality but suffers from quantization errors. The K-Means baseline is highly similar to the discretization baseline. However, instead of simply splitting targets into bins, we cluster the actions and optimize the policy with a classification loss to select the correct cluster \cite{pearce2023imitating}. The fourth baseline augments the K-Means with an observation-dependent residual used to modify the bin mean and trained with the MSE loss \cite{shafiullah2022behavior}. We also compare to a Gaussian action distribution parameterized by mean and standard deviation, and a generative energy-based model with derivative-free sampling \cite{florence2022implicit}.

We use the MLP sieve architecture for all learning-based methods and their implementations by \cite{pearce2023imitating}. Preliminary experiments with transformers were discarded due to limited computational resources, slow sampling, and no performance gain over the MLP sieve on NGSIM.

For parametric baselines, we use two variants of IDM \cite{treiber2000congested}. In the first variant with fixed parameters, we randomly sample one of the driving styles identified in prior work \cite{sunberg2017value}. The second variant involves using Optuna \cite{akiba2019optuna} to find a set of IDM parameters for each driver in the training data. We fit a multivariate Gaussian to the parameters and sample parameters from that during evaluation for the test set drivers.

\subsection{Results}

We evaluated our method, DSDP, and the chosen baselines in four NGSIM environments and highway-env using two primary metrics: the crash percentage (success rate) and the F1 score, which measures the similarity between generated and ground-truth driving trajectories. 

\textbf{Crash Percentages} (Table~\ref{tab:highway_env_crash_prob}): DSDP achieves the lowest crash percentages among fully learning-based methods in US 101, I-80, and Lankershim, with statistically significant improvements overall compared to other baselines. Our method outperforms all other methods in three environments, demonstrating robust generalization across diverse driving scenarios. The only exception is the Peachtree environment, where the crash rates across methods are within a statistically significant range due to the irregularity of the data, including frequent full stops due to traffic lights not modeled in our state space. While the IDM (Fixed and Learned) shows no crashes by construction, its hard-coded behavioral characteristics often lead to unhumanlike behavior, as our following evaluation shows. 

\textbf{F1 Scores} (Table~\ref{tab:multistep_f1}): Our method, DSDP, achieves the highest F1 scores overall and in three of the four tasks. The result highlights our method's ability to generate more human-like trajectories than the baselines, notably including the IDM variants, which fail to match the human-like driving styles despite learning the parameters that should provide it with that ability. The Peachtree environment remains challenging for all models due to the tabular state representation not fully capturing all the details of the environment.

\subsection{Ablations}

To understand which components are crucial to the performance of DSDP, we present a series of ablation studies analyzing the impact of various design choices. We perform all ablations with five seeds and report the results with two standard errors as uncertainty.

\begin{table}
\caption{Different sampling strategies.}
\centering
\begin{tabular}{l S[table-format=1.1] S[table-format=1.2] S[table-format=1.3] S[table-format=1.2]}
\hline
Method & \multicolumn{2}{c}{Mean Crash \%} & \multicolumn{2}{c}{Mean F1 Score} \\ \hline
DSDP & \textbf{4.0} & \pm0.79  & 0.404 & \pm0.01 \\
DSDP-KDE & 4.8 & \pm0.96  & 0.399 & \pm0.01 \\
DSDP-X (4) & 4.4 & \pm0.75  & \textbf{0.408} & \pm0.01 \\
DSDP-X (8) & 4.3 & \pm0.84  & 0.406 & \pm0.01 \\
DSDP-X (16) & 4.4 & \pm0.84  & 0.405 & \pm0.01 \\
DSDP-X (32) & 4.2 & \pm1.34  & 0.403 & \pm0.01 \\
\hline
\end{tabular}
\label{tab:sampling}
\end{table}

\begin{table}
\caption{Different architectural choices.}
\centering
\begin{tabular}{l S[table-format=1.1] S[table-format=1.2] S[table-format=1.3] S[table-format=1.2]}
\hline
Method & \multicolumn{2}{c}{Mean Crash \%} & \multicolumn{2}{c}{Mean F1 Score} \\ \hline
DSDP & \textbf{4.0} & \pm0.79  & 0.404 & \pm0.01 \\
w/ Uniform Prior & 7.2 & \pm3.81  & 0.372 & \pm0.01 \\
w/ VQVAE & 6.1 & \pm0.64  & 0.395 & \pm0.01 \\
w/ Action Loss & 5.1 & \pm0.99  & \textbf{0.414} & \pm0.01 \\
\hline
Diffusion-BC (baseline) & 6.3 & \pm0.31  & 0.376 & \pm0.00 \\
\hline
\end{tabular}
\label{tab:arch}
\end{table}

\begin{table}
\caption{Contrastive loss function.}
\centering
\begin{tabular}{l S[table-format=1.1] S[table-format=1.2] S[table-format=1.3] S[table-format=1.2]}
\hline
Loss function & \multicolumn{2}{c}{Mean Crash \%} & \multicolumn{2}{c}{Mean F1 Score} \\ \hline
InfoNCE & \textbf{4.0} & \pm0.79  & \textbf{0.404} & \pm0.01 \\
Triplet Loss & 4.8 & \pm0.72  & 0.393 & \pm0.01 \\
Barlow Twins & 7.5 & \pm2.33  & 0.370 & \pm0.01 \\
\hline
Diffusion-BC (baseline) & 6.3 & \pm0.31  & 0.376 & \pm0.00 \\
\hline
\end{tabular}
\label{tab:loss}
\end{table}

\begin{table}
\caption{Contrastive learning sub-trajectory length}
\centering
\begin{tabular}{l S[table-format=1.1] S[table-format=1.2] S[table-format=1.3] S[table-format=1.2]}
\hline
$L_c$ & \multicolumn{2}{c}{Mean Crash \%} & \multicolumn{2}{c}{Mean F1 Score} \\ \hline
5 & \textbf{4.0} & \pm0.79  & \textbf{0.404} & \pm0.01 \\
10 & 4.9 & \pm0.79  & 0.400 & \pm0.00 \\
20 & 4.7 & \pm0.73  & 0.398 & \pm0.00 \\
40 & 5.3 & \pm0.59  & 0.394 & \pm0.01 \\
\hline
Diffusion-BC (baseline) & 6.3 & \pm0.31  & 0.376 & \pm0.00 \\
\hline
\end{tabular}
\label{tab:len}
\end{table}

\begin{table}
\caption{Size of the LFQ codebook.}
\centering
\begin{tabular}{l S[table-format=1.1] S[table-format=1.2] S[table-format=1.3] S[table-format=1.2]}
\hline
Codebook Size & \multicolumn{2}{c}{Mean Crash \%} & \multicolumn{2}{c}{Mean F1 Score} \\ \hline
256 & \textBF{4.0} & \pm0.79  & 0.404 & \pm0.01 \\
128 & 4.9 & \pm1.26  & 0.404 & \pm0.01 \\
64 & 4.4 & \pm0.74  & \textBF{0.412} & \pm0.00 \\
32 & 5.1 & \pm0.76  & 0.402 & \pm0.01 \\
\hline
Diffusion-BC (baseline) & 6.3 & \pm0.31  & 0.376 & \pm0.00 \\
\hline
\end{tabular}
\label{tab:style}
\end{table}

Prior work has proposed replacing the default DDPM sampling with methods for selecting higher-likelihood actions \cite{pearce2023imitating}. Diffusion-X extends denoising by additional $X$ steps, and Diffusion-KDE samples multiple actions, fits a KDE, and selects the highest-likelihood action. We find that the differences between the methods are not statistically significant (Table~\ref{tab:sampling}). We test various architecture choices (Table~\ref{tab:arch}). Replacing the prior with a uniform prior disrupts style–observation alignment (leading to mismatched styles), and using VQVAE in place of LFQ yields poorer contrastive representations, both of which degrade overall performance. Adding an action reconstruction term during contrastive learning improves the F1 score but increases crash rates. The default contrastive loss in DSDP is InfoNCE, but we also evaluate Triplet Loss \cite{schultz2003learning} and Barlow Twins loss \cite{zbontar2021barlow}. Triplet loss is competitive with InfoNCE, whereas Barlow Twins fails to learn useful styles. Unlike InfoNCE, which leverages multiple negatives simultaneously, triplet loss has a single negative sample. Barlow Twins does not explicitly separate negatives and positives, making it less effective at distinguishing diverse styles. Finally, DSDP is robust to the sub-trajectory length and LFQ codebook size (Tables~\ref{tab:len},~\ref{tab:style}).

\section{Conclusion}

In this work, we propose the Discrete Style Diffusion Policy (DSDP) to model human driving behavior in the BC setting. DSDP combines unsupervised driving style learning from pre-existing datasets through contrastive methods, style discretization with Lookup-Free Quantization (LFQ), and a conditional Denoising Diffusion Probabilistic Model (DDPM) for action generation. Our method consistently outperforms other learning-based approaches in terms of safety (lower crash percentages) and human likeness (higher F1 scores) across multiple environments. Ablation studies reveal that DSDP is robust across different sampling strategies, sub-trajectory lengths, and codebook sizes, with the InfoNCE loss and learned prior being crucial components for capturing diverse driving styles.
\par
Future work on DSDP could explore several promising directions. First, simulating the limited observation capabilities of human drivers could lead to more realistic modeling of human-like decision-making processes. Second, scaling the method to learn from high-dimensional visual data would better mimic human perception during driving. Additionally, incorporating interactions with pedestrians and cyclists would enhance the model's fidelity to real-world driving scenarios. Although we applied discrete contrastive learning with DDPM in this work, other backbone methods could also be evaluated. Finally, exploring the application of DSDP's principles beyond the driving domain could open new avenues for modeling diverse human behaviors in other complex, interactive environments. 

\bibliographystyle{IEEEtran}
\bibliography{ref}

\end{document}